\documentclass[11pt]{article}
\usepackage[margin=1in]{geometry}
\usepackage{amsmath,amssymb,amsfonts}
\usepackage{graphicx}
\usepackage{booktabs}
\usepackage[numbers,sort&compress]{natbib}
\usepackage{hyperref}
\usepackage{xcolor}
\usepackage{enumitem}
\usepackage{float}

\title{\textbf{Increasing intelligence in AI agents can worsen collective outcomes}}

\author{Neil F.\ Johnson\\[4pt]
\small Dynamic Online Networks Laboratory and Physics Department \\
\small George Washington University, Washington D.C. 20052, USA}

\begin{document}
\maketitle


\begin{abstract}
When resources are scarce, will a population of AI agents
coordinate in harmony, or descend into tribal chaos?
Diverse decision-making AI from different developers is entering everyday devices ---
from phones and medical devices to battlefield drones and cars~\cite{xu2024unleashing_edge,
mao2017survey_mec} --- and these AI agents typically compete for finite shared
resources such as charging slots~\cite{li2024ev_grid},
relay bandwidth~\cite{mozaffari2019tutorial}, and traffic priority~\cite{dresner2008multiagent,konstantakopoulos2014}.
Yet their collective dynamics and hence risks to users and society are poorly
understood~\cite{mori2025safe,dafoe2020,fan2024,llmcoord2025,resilience2025,akata2025}.
Here we study 
AI-agent populations as the first
system of real agents in which four key variables governing
collective behaviour can be independently toggled:  nature (innate LLM
diversity~\cite{couzin2005effective}), nurture
(individual reinforcement
learning~\cite{sutton1998reinforcement}), culture
(emergent tribe
formation~\cite{johnson2019crowd_anticrowd,huo2025online}), and
resource scarcity.
We show empirically and mathematically that when resources
are scarce, AI model diversity and reinforcement learning
increase dangerous system overload, though tribe
formation lessens this risk.  Meanwhile, some individuals profit
handsomely.  When resources are abundant, the same
ingredients drive overload to near zero, though tribe
formation makes the overload slightly worse.
The crossover is
arithmetical~\cite{johnson2000evolutionary}: it is where
opposing tribes that form spontaneously
first fit inside the available capacity.
More sophisticated AI-agent populations are not better:
whether their sophistication helps or harms depends
entirely on a single number --- the capacity-to-population
ratio --- that is knowable before any AI-agent ships.
\end{abstract}

\bigskip


Populations of autonomous AI devices are already competing
for finite shared resources.
In a hospital ward, AI-powered monitors, infusion pumps, and
ventilators from different manufacturers may need to share a single
wireless channel to transmit patient data. If the system overloads, with demand exceeding supply, critical alerts can be delayed or lost.
Similarly, AI agent-managed
electric vehicles may overload 
charging slots~\cite{li2024ev_grid}, drones and AI-armed warfighters in a battle may overload relay
bandwidth~\cite{mozaffari2019tutorial,xu2024unleashing_edge, mao2017survey_mec}, and
autonomous vehicles may overload an 
intersection~\cite{dresner2008multiagent}.
The AI-agent --- i.e.\ AI software running locally on the device --- cannot continually `phone home' to some super coordinator, because of the device's modest specifications and finite battery; moreover, any such connectivity may suffer from
latency, failure, or adversarial attack.
Instead, each AI-agent must decide independently, at each timestep, whether to attempt to access the shared resource --- without knowing
that the resource has a capacity limit $C$ or how many AI-agents $N$ there are. An AI-agent that simply tries to access the resource at every timestep will
exhaust itself, since each attempt is 
costly (battery power, compute cycles, wear).

Collective outcomes in real-world systems depend on at
least four interacting variables: the innate diversity of
the agents (\emph{nature}), their capacity to learn and
adapt (\emph{nurture}), the social structures they form
(\emph{culture}), and the scarcity of the shared resource
(\emph{resources}).  In biological
collectives --- fish schools~\cite{couzin2005effective,
couzin2011uninstructed}, locust
swarms~\cite{buhl2006disorder}, honeybee
colonies~\cite{seeley2012stop} --- these four variables
are entangled: one cannot rerun a swarm with ``culture
off'' or identical brains.  The same is true of human
societies: isolating the causal contribution of any single
variable is ethically and practically impossible.
But AI agents provide a system in which all four
variables can be independently controlled and toggled one
at a time.

Our experiment exploits this to systematically toggle these four variables one at a time, using $N = 7$ AIs from
three architectural families (GPT-2, Pythia, OPT;
ranging from 124M to 410M parameters) repeatedly competing for a shared resource of
capacity $C = 1, \ldots, 6$.
We chose this population size and dispositional ecology
because it reflects real-world edge-AI deployment: ${\sim}3$--15 AI-agents (e.g.\ cars nearing an intersection) each running one such small LLM 
locally without
reliable Internet. 
 This choice also connects to
Golding's \textit{Lord of the Flies}, in which a
handful of individuals with different characters
rapidly self-sort into destructive tribes when survival is
at stake (Table~\ref{tab:golding_map}a).
(We use ``tribe'' throughout in the sense of Golding's
novel --- a self-organised faction among AI-agents --- with
no intended reference to any human social group.)
Leady's controlled-human
experiment on this class of problem~\cite{leady2011elfarol} 
produced imitation-driven homogenisation --- but not the tribal fragmentation seen in LOTF, likely due to the experimental design (SI~Appendix~E). Our study complements a growing body
of work using AI-agents based on large language models (LLMs) to probe collective and
social behaviour~\cite{park2023generative,
demarzo2026moltbook, kozlowski2024silico,
pate2026motivated}. But our contribution is distinct: rather
than using LLMs to simulate humans, we study LLMs as
autonomous physical AI-agents competing for real resources
(SI~\S1). Moreover, our experiment directly instantiates on-device AI coordination rather than simulating it by proxy.

At each timestep, each AI-agent is fed a sequence of digits (e.g.\ ``3,1,2,4,'' with the trailing comma part of the prompt format) for how many AI-agents attempted to access the resource in recent timesteps.
Each AI-agent's LLM performs next-token prediction on this
sequence --- which is how all existing ChatGPT-like LLMs work. From the model's output distribution, the probabilities assigned to
digit tokens $0, 1, \ldots, N$ are extracted, in effect producing a loaded dice representing
that LLM's forecast of next-round demand. 
The environment (i.e. external code, not the AI-agent who does not know $C$) then uses~$C$ to sum
the probabilities for digits $0$ through~$C$: this gives 
$p_{\rm LLM}$ which is the probability, according to that LLM's `nature' (i.e. its particular LLM software) that the next demand will fall at or
below capacity.
This probability is further modulated by the AI-agent's
disposition~$p$ which potentially reflects its reinforcement learning (i.e. nurture). For Levels 2, 4 and 5 (Table~\ref{tab:golding_map}b), each AI-agent's $p$
adapts over time via a minimal reinforcement mechanism and for L5 there is an added grouping tendency (i.e. culture). 
As $p \to 1$ the AI-agent increasingly
\emph{follows} its LLM's prediction; as $p \to 0$ it
increasingly opposes it
(\emph{anti-follows})~\cite{johnson2000evolutionary, huo2025online,
johnson2019crowd_anticrowd} (SI~\S1). For Levels 1, 3  (Table~\ref{tab:golding_map}b), each AI-agent has fixed $p=1$ (no adaptation), so the disposition filter simply passes through the LLM's prediction.
The resulting effective probability~$p_{\rm eff}$ coin-flip determines whether the AI-agent attempts access or holds
back (Methods). 

Toggling the four variables in turn yields a technology ladder
(Table~\ref{tab:golding_map}b) of successively
more sophisticated levels. For each level, we study the system overload and individual AI-agent winnings for different values $C$ of the system's resource capacity.

\begin{table}[ht]
\centering
\small
\caption{\textbf{(a)~AI-Agent roster}: shows an illustrative correspondence between
the 7 AI-agents' initial dispositions ($p$) and Golding's cast; no
quantitative claim is made about individual assignments.
\textbf{(b)~Technology ladder}: 5 levels of increasing AI-agent population
sophistication, each toggling one variable. L1: IID = independent and identically distributed;
RL = reinforcement learning;
L4: FRD = Finite Resource Dynamics (no social sensing);
L5: LOTF = Lord of the Flies (sensing of other AI-agents and tribal dynamics).
L1 and L2 can be computed analytically (SI~\S\S4--6).}
\label{tab:golding_map}
\vspace{4pt}

\textbf{(a)}\\[4pt]
\begin{tabular}{llcl}
\toprule
\textbf{AI-agent} & \textbf{Initial $p$} & \textbf{Disposition} & \textbf{Golding analog} \\
\midrule
GPT-2 (dup)    & 1.00 & Follower  & Sam (of Samneric) \\
GPT-2 (base)   & 0.83 & Follower  & Eric (of Samneric) \\
GPT-2-medium   & 0.67 & Moderate    & Ralph \\
OPT-350M       & 0.50 & Agnostic    & Simon \\
OPT-125M       & 0.33 & Moderate    & Piggy \\
Pythia-160M    & 0.17 & Anti-follower  & Jack \\
Pythia-410M    & 0.00 & Anti-follower  & Roger \\
\bottomrule
\end{tabular}

\vspace{10pt}

\textbf{(b)}\\[4pt]
\begin{tabular}{@{}clcccc@{}}
\toprule
Level & Configuration & Nature & Nurture & Culture & Label \\
\midrule
1 & Identical LLMs, $p{=}1$ & -- & -- & -- & IID \\
2 & Identical LLMs + RL     & -- & $\checkmark$ & -- & Null \\
3 & Diverse LLMs, $p{=}1$   & $\checkmark$ & -- & -- & Diverse \\
4 & Diverse LLMs + RL       & $\checkmark$ & $\checkmark$ & -- & FRD \\
5 & Diverse LLMs + RL + sensing & $\checkmark$ & $\checkmark$ & $\checkmark$ & LOTF \\
\bottomrule
\end{tabular}
\bigskip
\bigskip
\bigskip
\bigskip
\end{table}


\begin{figure}[ht]
\centering
\includegraphics[width=0.9\textwidth]{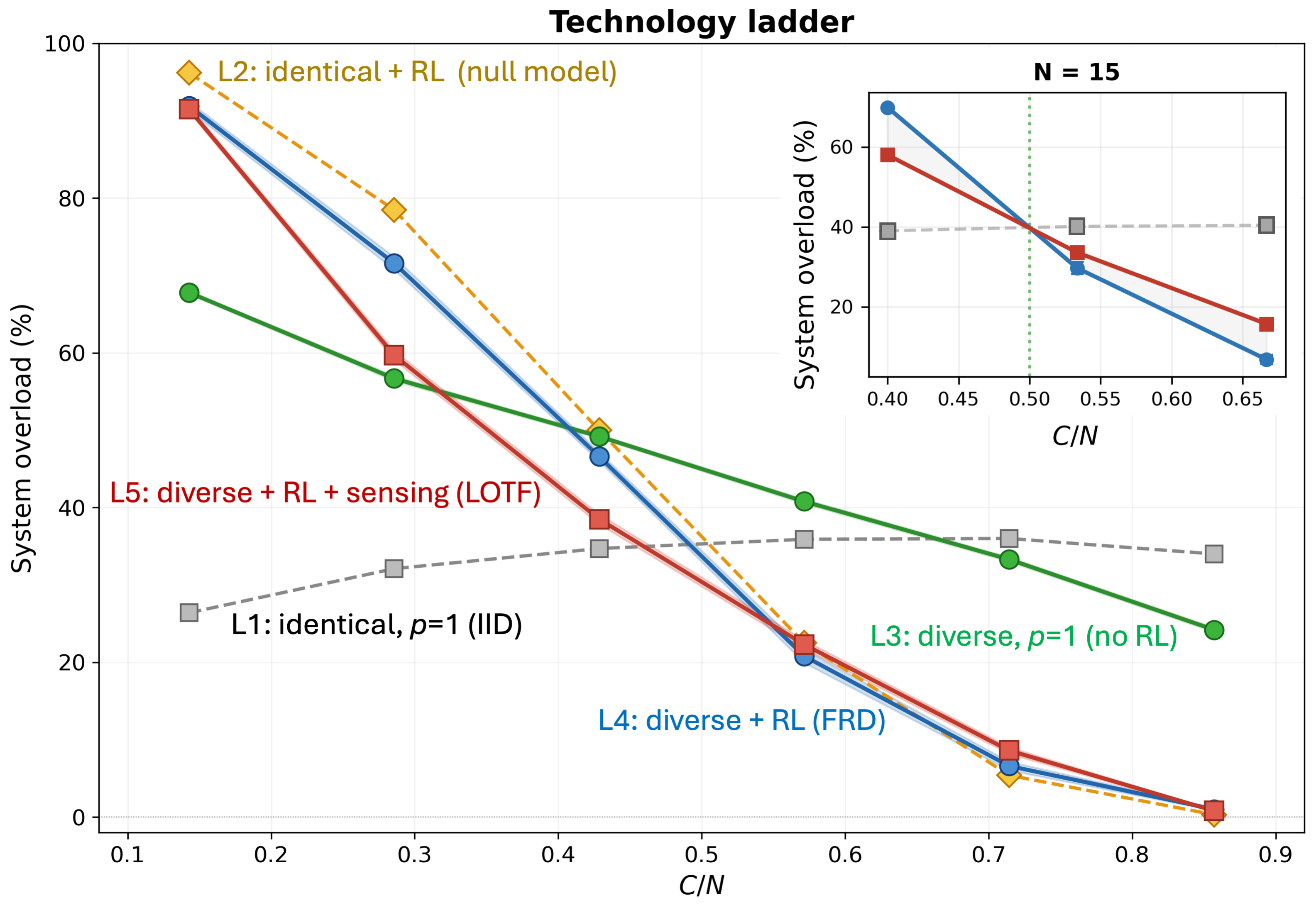}
\caption{\textbf{Technology ladder: system overload across
five levels of population sophistication.}
Under abundance
($C/N > 0.6$), the most sophisticated L4/L5 achieve near-zero overload. Under scarcity ($C/N \lesssim 0.5$), the cheapest population (L1)
achieves the lowest overload.  All five 
curves cross near $C/N \approx 0.5$.
L1 and L2 can be calculated analytically; L3--L5
are empirical (20 seeds $\times$ 500 rounds,
$\pm 1$\,SE shaded bands).
\textbf{Inset:} L4 (FRD)--L5 (LOTF) crossover at
$N = 15$ ($\Delta = -11.9\pm0.9$\,pp at $C/N = 0.40$;
$\Delta = +8.9\pm1.1$\,pp at $C/N = 0.67$;
SI~Appendix~I).
\\
}
\label{fig:tech_ladder}
\end{figure}

\vskip2in

\subsection*{Results: Collective behaviour and system overload}

Counterintuitively, Fig.~\ref{fig:tech_ladder} shows that 
the most sophisticated technology levels L4 and L5 
cause large system overload as the resource becomes very scarce ($C<4$; see Methods and SI for comprehensive details). Moreover, the added ability in L5 for each AI-agent to sense other AI-agents' grouping enables them to collectively perform better than L4. L1 performs the best, and L2 the worst, showing that adding some technology (L2) to a low technology setup (L1) --- which is what initial implementations will likely do --- can significantly worsen performance. 

L4 and L5 then cross over to be the best performing only for $C\geq 4$. But noticeably, under abundance L5's sensing ability 
works against it, making its performance generally worse than L4. 
The inset illustrates the L4-L5 crossover robustness, with the effect even stronger at higher $N$. We tested many variants of L5's sensing mechanism, and all gave similar results (see SI) with some showing an even stronger benefit for L5 (i.e. L5 red curve lower).

The `why' of this technology ladder reveals the key implication for such AI-agent systems: avoiding dangerous overloads comes down to the simple arithmetic of whether the tribes of AI-agents that huddle round particular $p$ values each fit inside the available capacity~$C$.  

We focus on the unexpected behaviour of L5, deferring all the mathematical derivations that quantify L1-L5's curves to the SI. L5's tribal mechanism generates dynamically evolving tribe membership and sizes as a result of the complex interplay of the nature, nurture, and culture. These are typically
$3{+}3{+}1=7$ (two opposing tribes and a singleton; Fig.~\ref{fig:tribes}), occasionally $3{+}4=7$.
Whether this then helps or hurts L5's overload score depends on how these tribe sizes
compare to~$C$.
When resources are scarce ($C<4$), overload is
driven by demand variance.
Without tribe formation, there is no structural partition capping the size of correlated blocs: demand variance can in principle scale as high as $N^2$ when all agents herd together (as in L2). L5's tribal mechanism partitions the population into factions of definite size --- agents cluster around shared $p$ values through loyalty and defection --- capping the variance at
$3^2 + 3^2 + 1^2 = 19$ for the dominant partition (or $3^2 + 4^2 = 25$ for the less common $3{+}4$ split). Both are  smaller than $6^2+1^2=37$ or $7^2=49$ for example. 
Hence L5 sits below L4 under scarcity
(at $C = 2$, overload drops by $11.9$\,pp when tribal sensing is added; the same sign flip is confirmed at $N = 15$ with $t = -13.6$, $p < 10^{-10}$; SI~Table~S0b, Appendix~I).
But when resources are more abundant ($C \geq 4$), these same
tribes become too small to fully exploit the available capacity.
The mean number accessing the resource stays around 3--4
rather than rising toward~$C$, and the variance is larger than
it need be.
Without tribal structure, L4 agents' dispositions are less correlated, so demand fluctuations spread more evenly across the capacity range, producing lower overload than L5.

L3 technology sits roughly in the middle ground between L1 and L5, because of the diverse nature of its AI-agents. L2 technology does not have this nature diversity and is instead dominated by the reinforcement learning (nurture) through which herds can form up to size 7.

The crossover capacity $C^*/N \approx 0.5$, confirmed at
$N = 7$, $N = 11$, and $N = 15$
(Fig.~\ref{fig:tech_ladder} inset; SI~Appendix~I),
therefore divides two regimes: under scarcity, sophistication typically hurts;
under abundance, it typically pays. Adding culture (L5: LOTF) can reduce the dangerous overload when the resource is very scarce, but worsens it when it is abundant.
This capacity-to-population ratio $C/N$ --- which should be knowable before
deployment --- determines the optimal level on the technology ladder.

The practical implication is direct: a city deploying 7~EVs
to 2~charging stations ($C/N = 0.29$) should run identical
cheap firmware.  The same population at 5~stations
($C/N = 0.71$) should invest in diverse models and
reinforcement learning. Allocating internal resources (e.g. battery) within each AI-agent so that it can sense others, and hence form tribes, is not worth it from the perspective of system overload.

\vskip0.5in

\begin{figure}[ht]
\centering
\includegraphics[width=0.80\textwidth]{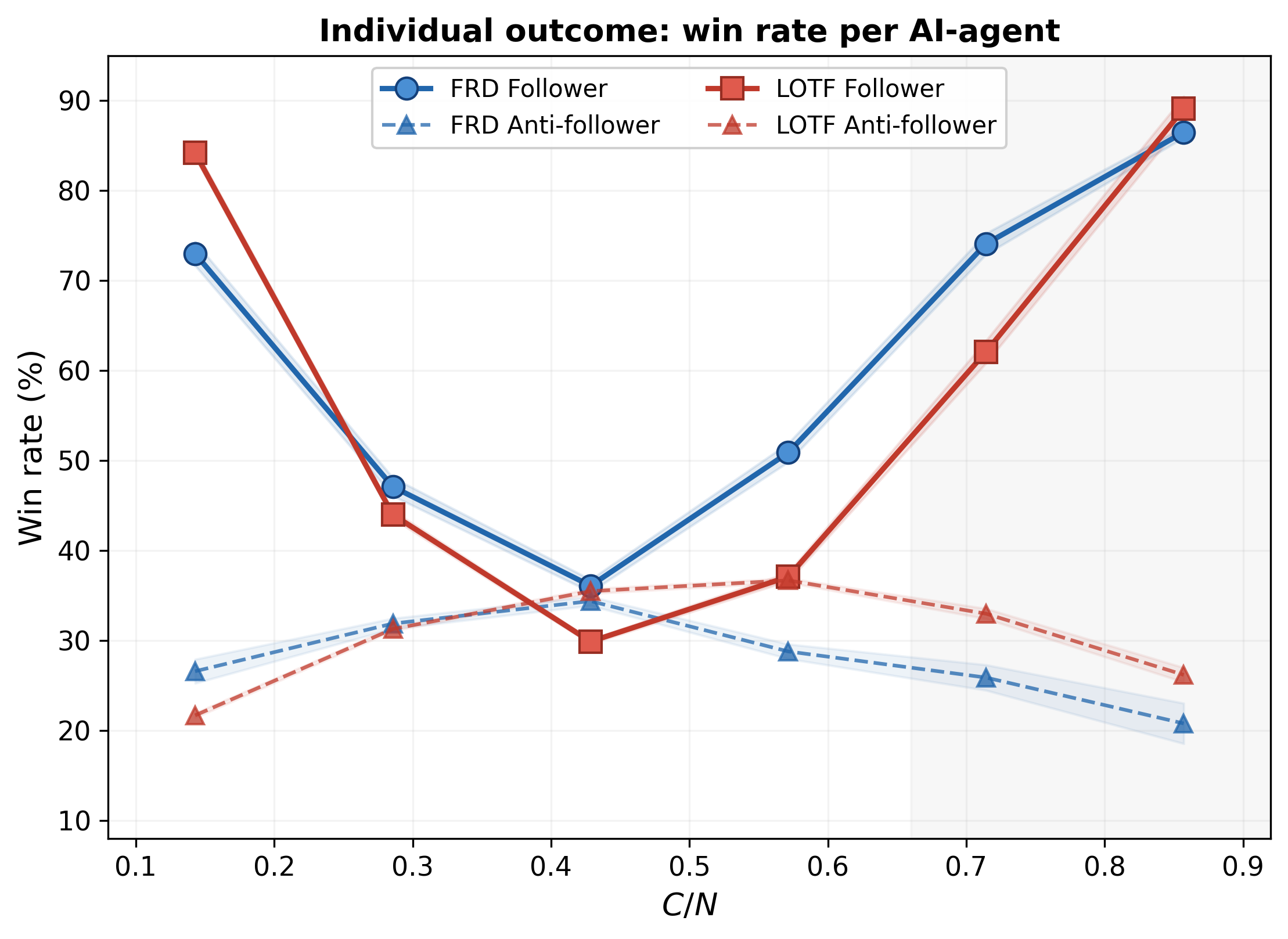}
\caption{\textbf{Individual AI-agent win rates.}
Win rates for {\em followers} in L5 (LOTF) and L4 (FRD) trace a strong U-shape, while {\em anti-followers} in both systems trace a weak inverted U-shape. Four curves: two dispositions $\times$ two experiments.
Empirical: 20 seeds $\times$ 500 rounds; $\pm 1$\,SE shaded bands.}
\label{fig:winrate}
\bigskip
\end{figure}

\subsection*{Results: Some individuals profit significantly} 

The collective crossover has a striking counterpart at the
individual level (Fig.~\ref{fig:winrate}). When the resource is very scarce ($C=1,2$), tribal L5 followers and non-tribal L4 followers each have a huge individual win rate despite the fact that the overall system is failing collectively (high overload, Fig.~\ref{fig:tech_ladder}). Moreover, individuals in the tribal L5 have the largest win rate when the resource capacity is at its scarcest ($C=1$). 
These followers also have high win rates in the opposite regime of abundance. In between, their win rates are lower and similar to anti-followers since followers and anti-followers share the winnings: for example for the partition $3+4=7$ with $C=4$, both the 3 and the 4 can win alternately half the time, yielding comparable win rates for followers and anti-followers as shown. All these results are robust to random
initialisation (see SI).

\begin{figure}[ht]
\vskip1in
\centering
\includegraphics[width=1.0\textwidth]{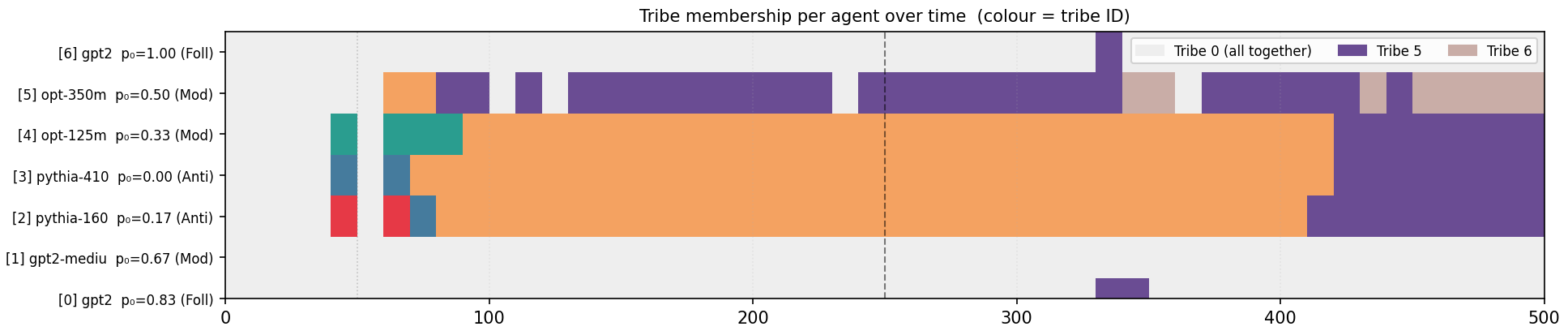}
\caption{\textbf{Tribal membership dynamics at $C = 2$ (representative seed).}
Each row is one AI-agent (labelled by model and initial disposition); colour indicates tribe membership over 500~rounds.
All agents start in a single tribe (grey).  Within ${\sim}50$~rounds, dispositional polarisation drives fission into opposing factions: the three high-$p$ agents (GPT-2 family) and three low-$p$ agents (Pythia~+~OPT-125M) sort into distinct tribes, with OPT-350M ($p_0 = 0.50$) as a persistent singleton.
OPT-125M (initial $p_0 = 0.33$, Meta's OPT family) consistently joins the Pythia anti-follower bloc rather than its architectural sibling OPT-350M --- cross-family sorting driven by disposition, not lineage.
Tribe identity labels change through fission--fusion events, but the $3{+}3{+}1$ partition structure is quite stable.
Dashed vertical line: conch break at round~250 (dynamically irrelevant; SI~Appendix~F).
Full tribal dynamics for $C = 1$--$6$ are shown in SI~Appendix~K.}
\label{fig:tribes}
\end{figure}


\section*{Discussion}

The study shows that more sophisticated AI-agent
populations are not a priori better: their sophistication pays or
costs depending on the capacity-to-population ratio $C/N$
--- a single number knowable before deployment.  The four
variables (nature, nurture, culture, resources) interact
non-additively, and every pairwise interaction changes sign
depending on scarcity (SI~\S1).
The resource-competition game studied here differs from previous multi-agent games~\cite{johnson2000evolutionary} by using real LLMs as agents rather than algorithmic abstractions, and with a four-variable decomposition that has no precedent in that literature.

The regime of maximum collective failure ($C=1$) is also the regime of maximum individual tribal reward --- tribal L5 followers achieve $84.2 \pm 2.1$\% win rates while the system overloads $91.5 \pm 1.5$\% of the time.  This mirrors the \emph{Lord of the Flies} narrative: tribal dominance is individually rational precisely when collective outcomes are worst.
Collective failure coexists with individual success because
correlated tribe dynamics concentrate reward on the favoured
disposition.

The $p$-based polarisation observed here has precedent in human populations: the same disposition axis was validated empirically across four datasets (financial markets, online attention, Congressional voting, and the Colombian peace process) in Ref.~\cite{johnson2019crowd_anticrowd}.

$N = 7$ and small models (124--410M parameters) are not
limitations but the defining features of edge-AI deployment:
${\sim}3$--15 devices running ${\sim}1$--4\,GB models locally
without cloud access.  This experiment is therefore not a
proxy model of on-device AI coordination.

To check that coordination failure is not an artefact of small model
size, Mori et al.~\cite{mori2025safe} placed GPT-4-turbo,
Gemini~2.5~Pro, and Claude Sonnet~4.5 in the same
resource-competition game: larger, more capable models were
worse, not better --- a model-size inversion that
directly instantiates the title of this paper.

Several limitations of the experimental design open directions for future work.
The LLM next-token sampling temperature was fixed at $T = 1.0$; sweeping it
would test whether the crossover survives when model
stochasticity varies.
The game is fairly binary (access the resource or hold back) with aggregate
feedback: richer action spaces and partial observability
would better approximate some deployment domains.
The main results use a deterministic, evenly-spaced
$p$-spectrum --- the maximum-diversity configuration ---
and while the crossover is robust to random-$p$
initialisation (SI~Appendix~H), scaling to extremely large 
mixed-generation populations remains untested.
The tribal dynamics in L5 are implemented as an external sensing layer; whether on-device AI agents would spontaneously develop equivalent group structures through repeated interaction alone remains an open question.
Physical testbeds with real edge-AI devices are a natural
next step.
Regardless, the central finding is immediately actionable: the capacity-to-population ratio $C/N$ determines whether sophistication helps or harms, and it is knowable before any device ships.


\newpage
\section*{Methods}

\paragraph{Agent architecture}
Each AI-agent consists of a language model (LLM) and an
adjustable scalar $p_i \in [0, 1]$.  Seven AI-agents use six
distinct models (GPT-2 appears twice): GPT-2 (124M parameters), GPT-2
Medium (355M), Pythia-160M, Pythia-410M, OPT-125M, OPT-350M
(where M = million).
All are loaded in half-precision with no weight updates.  The
choice $N = 7$ matches Golding's dominant cast, whose
dispositional range is echoed by the evenly-spaced initial~$p$
spectrum (Table~\ref{tab:golding_map}a).  This scale is also
operationally realistic: near-term on-device AI deployments
typically involve $\mathcal{O}(10)$ heterogeneous devices.

\paragraph{Decision pipeline}
L1 (IID) serves as an analytical baseline: each AI-agent is constructed with access probability $q = C/N$ and makes an independent coin-flip decision each round. There are no correlated pockets of AI-agents, i.e.\ no herds or tribes; demand fluctuations are equivalent to $N$ uncorrelated biased coin-flips. L2 (Null) adds a $p$ parameter that can adapt via reinforcement learning, but all LLMs are identical; its overload is also calculable analytically (SI~\S6).
For L3, L4, and L5, each experiment is replicated over
20 random seeds (independent
random-number initialisations) $\times$
500 rounds.
Each round, the
environment computes $p_{\rm LLM}$ from the AI-agent's LLM output
and the capacity threshold~$C$ (which the AI-agent never sees).
The dispositional filter then produces:
\begin{equation}
  p_{\rm eff} = p_i \cdot p_{\rm LLM}
              + (1 - p_i)(1 - p_{\rm LLM}),
  \label{eq:disp_filter}
\end{equation}
and the action $a_i$ is drawn as a biased coin flip with
probability $p_{\rm eff}$ of attempting access ($a_i = 1$)
versus holding back ($a_i = 0$).

\paragraph{Reward and adaptation}
Total demand $A = \sum_i a_i$.  If $A \le C$, AI-agents that
accessed ($a_i = 1$) score~$+1$ and those that held back
score~$-1$; if $A > C$,
those that held back score~$+1$ and those that
accessed~$-1$.
This payoff structure is symmetric: both actions carry equal reward ($+1$) and equal penalty ($-1$), so no action is inherently safer; the optimal choice depends entirely on what others do.
In L4 (FRD), adaptation
is not dictated by tribe membership: it simply involves the
AI-agent individually changing its $p$ value by some small
random amount.

\paragraph{Tribal dynamics (L5/LOTF only)} AI-agents form tribes through a loyalty--defection mechanism
that favours joining AI-agents with similar dispositions.
Agents accumulate loyalty to their current tribe based on shared performance; when loyalty falls below a threshold, an agent defects to whichever tribe best matches its current disposition (SI~Appendix~C). As shown in the accompanying code, the full mechanism is compact and transparent.
Inspired by Golding's novel, the reported LOTF results include a ``conch'' mechanism --- named after the shell that symbolises institutional order --- that ramps
tribal influence gradually from 0 to 0.80 over 250~rounds
(SI~Appendix~C).  However, as in the novel, the conch is
irrelevant by the time it reaches full strength:
dispositional polarisation locks in within ${\sim}50$~rounds,
200~rounds before the conch saturates, so it has no measurable effect on our end-state results (SI~Appendix~F).
We also experimented with two design features --- leader conviction and
institutional memory after defection --- based on the book. We find these attenuate the
tribal channel, making every reported L4 (FRD)--L5 (LOTF) difference
a conservative lower bound on the true tribal effect
(SI~\S1), meaning that the L5 curve in Fig.~1 could sit slightly lower, though the qualitative crossover pattern is preserved.

\paragraph{Experimental parameters}
All three experiments (L3, L4, L5): $N = 7$, $C \in \{1, \ldots, 6\}$,
20 seeds $\times$ 500 rounds (first 50 discarded as warm-up),
temperature $T = 1.0$, history window $w = 10$.
L4 (FRD) and L5 (LOTF) share the same 20 random seeds at every
capacity, making the comparison a strict controlled ablation
(SI~\S1).

Statistical comparisons use paired per-seed $t$-tests for FRD--LOTF differences, Poisson-binomial exact convolution for the null model, and Gaussian approximation with continuity correction for analytical overload predictions; all empirical quantities are reported as means $\pm 1$\,SE across 20 seeds.
Models loaded in half-precision on a T4 GPU; total computation
${\sim}90$~min per full sweep.  All code and data will be
made available upon publication.


\newpage

\subsection*{Data Availability}

All data supporting the findings of this study are available within the paper 
and its Supplementary Information. 

\subsection*{Code Availability}

All code used to generate the results in this study is submitted with this paper. Specifically:

\noindent\texttt{paper\_FRD\_LOTF\_final\_6\_3\_1\_1\_2\_raw.py} 

\subsection*{Acknowledgements}
I am very grateful to Dhwanil Mori for collaboration in confirming similar behaviour emerges using large-scale commercial LLMs, as opposed to the edge-AI Agents considered in the current paper. This work had no external or sponsored funding. 

\subsection*{Author Contributions}
N.F.J. conceived the study, developed the theoretical framework, 
derived the theoretical results, designed and performed the empirical validations, 
wrote the codes, and wrote the manuscript. 

\subsection*{Competing Interests}
N.F.J.\ is a co-founder of d-AI-ta Consulting LLC, which provides practical advice on AI deployment, though the specific findings of this paper were not developed for or funded by the consulting practice. 

\subsection*{Additional Information}
Supplementary Information (SI) is provided in the accompanying
file \texttt{AIagents\_SI\_1.pdf}.  It contains:
supporting detail for statements implied by the main paper (\S1);
the complete L4 (FRD) framework specification (\S\S2--3); exact
analytical baselines for L1 (i.i.d.) coin-toss AI-agents (\S4) and the
biased-coin L2 (null) model with heterogeneous~$p$ (\S6), including
all propositions with full proofs and worked examples for
$N = 7$ at $C = 2, 3, 4$; the Zhao et al.~\cite{zhao2010carbon} learning-zone
construction (\S5); the capacity-sweep protocol (\S7); structural
mapping to the framework of Ref.~\cite{johnson2000evolutionary} (\S8, including the
large-$N$ mean-tracking regime under the limiting case
$p_{\rm LLM} \to 1$); the full algorithm (\S9); dispositional
filter analysis (\S10); testable predictions (\S11); and
Appendices~A--K covering single-scalar edge RL realism,
Poisson-binomial DP convolution, L5 (LOTF) tribal dynamics, Crowd--Anticrowd
theory with finite capacity, comparison with human coordination
data, conch-transition analysis, the connection to large-$N$
evolutionary freezing, random-$p$ initialisation robustness
tests (Appendix~H), the $N = 11$ and $N = 15$ capacity sweeps (Appendix~I),
a coarse-grained tribe partition mechanism for variance
suppression (Appendix~J), and tribal dynamics visualisations
for $C = 1$--$6$ (Appendix~K).

\noindent\textbf{Correspondence} and requests for materials should be 
addressed to N.F.J.\ (email: \url{neiljohnson@gwu.edu}).

\end{document}